\documentclass{article}

\usepackage{spconf,amsmath,epsfig}
\usepackage{times}
\usepackage{graphicx}
\usepackage{amssymb}
\usepackage{enumitem}

\title{Single Source One Shot Reenactment using \\Weighted Motion from Paired Feature Points}

\twoauthors
  {Soumya Tripathy\sthanks{Corresponding author}, Esa Rahtu
  }
  {Computer Vision Group \\ Tampere University\\
  Tampere, Finland\\
  {soumya.tripathy, esa.rahtu}@tuni.fi}
  {Juho Kannala}
  {Department of Computer Science \\Aalto University\\
  Helsinki, Finland\\
  juho.kaanala@aalto.fi}

\begin{document}

\maketitle
\sloppy
\begin{abstract}

Image reenactment is a task where the target object in the source image imitates the motion represented in the driving image. One of the most common reenactment tasks is face image animation. The major challenge in the current face reenactment approaches is to distinguish between facial motion and identity. For this reason, the previous models struggle to produce high-quality animations if the driving and source identities are different (cross-person reenactment). We propose a new (face) reenactment model that learns shape-independent motion features in a self-supervised setup. The motion is represented using a set of paired feature points extracted from the source and driving images simultaneously. The model is generalised to multiple reenactment tasks including faces and non-face objects using only a single source image. The extensive experiments show that the model faithfully transfers the driving motion to the source while retaining the source identity intact.


\end{abstract}

\section{Introduction} \label{Introduction}
General image reenactment and particularly facial reenactment have received plenty of attention in recent years due to numerous applications in game design, movie
production, virtual reality, and interactive system
design. 
 The current
state of the art models \cite{siarohinFirstOrderMotion2019,
haMarioNETteFewshotFace2019, zakharovFewShotAdversarialLearning2019,
burkovNeuralHeadReenactment2020, yaoMeshGuidedOneshot2020,
tripathyFACEGANFacialAttribute2020, nirkinFSGANSubjectAgnostic2019}
can produce realistic talking heads of a source from a single
image by imitating the facial movements from another similar looking talking
video, commonly known as the driver. Impressive results often require careful selection of source and driving pairs with closely matching identities. For example, models like \cite{zakharovFewShotAdversarialLearning2019,
nirkinFSGANSubjectAgnostic2019, wilesX2FaceNetworkControlling2018,
zakharovFastBilayerNeural2020a} generate high-quality talking heads
for a person who drives his own face (self-reenactment) or a face with a comparable head
structure. Other models \cite{yaoMeshGuidedOneshot2020,
tripathyICfaceInterpretableControllable2020,
tripathyFACEGANFacialAttribute2020,
haMarioNETteFewshotFace2019, thiesFace2FaceRealtimeFace}
require facial identity and motion representations in terms of 3D
models or pretrained representation like landmarks, head poses or
Action Units (AUs) \cite{ekmanFacialActionCoding1978}. These pretrained
models usually require costly annotations and often fail to handle
occlusions or extreme head poses.

\begin{figure*}[t]
  \centering
  \includegraphics[scale=0.82]{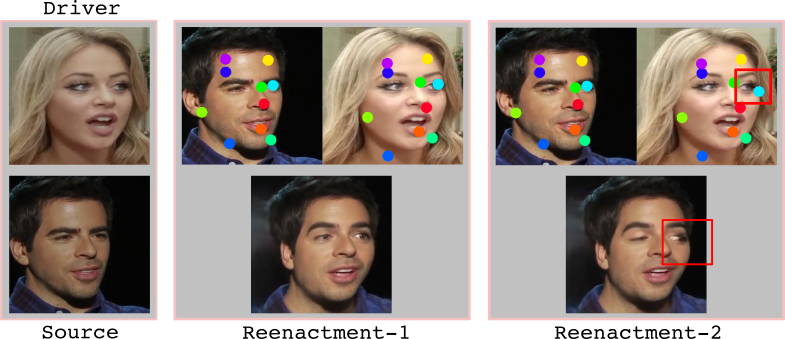}
  \caption{\label{fig:keypoint_drawbacks} Illustration of drawbacks in
    keypoints/landmarks based reenactment models. In both
    cases, the reenactment is performed using FOM
    \cite{siarohinFirstOrderMotion2019} and the keypoints are drawn on the source and driving images. In Reenactment-1, the head structure difference between the source and driving is reflected in the output (bottom image) as the source's facial structure and identity are distorted. In the Reenactment-2, one of the key points
    (in the red box) is slightly displaced manually from its original position to show its effect on the output. The degradation in the output quality shows the overall system performance is highly dependant on the keypoint detectors.}
\end{figure*}

Some of these issues are tackled in
\cite{siarohinAnimatingArbitraryObjects2019,
  siarohinFirstOrderMotion2019} using unsupervisedly learned motion representation defined as a function of key points. Although the keypoint detector is obtained without explicit annotations, the learning process is driven by
objective functions like equivariance loss, which encourage landmark like locations (e.g. lip corners). Figure \ref{fig:keypoint_drawbacks} illustrates examples of detected keypoints \cite{siarohinFirstOrderMotion2019}, which are mostly located on facial contours.
The design choice is reasonable as most of the state of
the art reenactment models \cite{zakharovFewShotAdversarialLearning2019, haMarioNETteFewshotFace2019, nirkinFSGANSubjectAgnostic2019,
tripathyFACEGANFacialAttribute2020, wuReenactGANLearningReenact2018}
use landmark-based motion representations. However, the landmark and contour-driven locations are prone to contain substantial shape information. One important advantage in the unsupervisedly learned keypoints is the fact that they are object agnostic and can be used to animate other objects than faces.

Face landmarks or keypoint based models generate high-quality talking
heads for self reenactment, but often fail in cross-person reenactment where
the source and driving image have different identities.
The main reason is that landmarks are person-specific and carry facial shape information
in terms of pose independent head geometry \cite{burkovNeuralHeadReenactment2020,
tripathyFACEGANFacialAttribute2020}. Any differences of shape between
source and driving heads are reflected in the facial motion (through landmarks or
keypoints) and lead to a talking head that can not faithfully retain the identity
of the source's person. This effect can be seen in Figure \ref{fig:keypoint_drawbacks}
for faces and in Figure \ref{fig:mgif}
for non-face objects using a keypoint based reenactment model \cite{siarohinFirstOrderMotion2019}. Furthermore,
these models use each keypoint independently
to affect the motion of its neighborhood pixels which makes the output highly dependent
on the quality of the keypoints or landmarks. Any noisy keypoint prediction may severely
distort the facial shape and thereby generate low-quality talking heads of
the source as shown in Figure \ref{fig:keypoint_drawbacks}.

Considering the aforementioned issues in the existing reenactment models, we propose
two important improvements. First, we propose a new paired feature point detector that predicts anchors on the source and driving images that
best describe the motion between them without imitating the
landmarks.
Due to lack of the physical significance like landmarks, we refer to the detected locations as feature points instead of keypoints. In previous works \cite{siarohinFirstOrderMotion2019,zakharovFewShotAdversarialLearning2019,
  nirkinFSGANSubjectAgnostic2019}, the keypoint predictions (supervised and self-supervised) are extracted independently from the source and driving images. Such setup prevents optimising the locations for a specific source and driving identities leading to landmark like keypoints. In contrast, we predict the feature points using source and driving images jointly using a multi-headed co-attention layer. Hence, we call them as paired-feature-points
instead of keypoints throughout the paper.
The paired-feature-points encourage the detector to
predict different features for different pairs without encoding the
facial shapes.

Second, we propose a new motion model that
predicts the motion for each source pixel using all the paired-feature-points. Here we use the Moving Least Square
\cite{schaeferImageDeformationUsing} formulation where each pixel's
motion depends on all the paired-feature-points, unlike the
first-order-model \cite{siarohinFirstOrderMotion2019} where
each corresponding keypoint is only responsible for the motion of
its neighborhood pixels. Our model is less dependent on the
correctness of any individual keypoint and unlikely to fail in
conditions where some of the keypoints' positions are wrong due to
the occlusions or changing head poses. We use a simple and
robust formulation to express the motion with paired feature points. Our model is applicable to both face and non-face objects similarly to \cite{siarohinFirstOrderMotion2019, siarohinAnimatingArbitraryObjects2019}.


We show that our paired-feature-point detector and motion model can
be used to effectively reenact a face from a single image without any strong priors on
the identities, initial pose, or representations (like landmarks,
action units) unlike any other state-of-the-art model. In the facial
cross-person reenactment, we show experimentally that our model
preserves the identity better than other one-shot
reenactment models. In addition, we compare the proposed model with few-shot learning based models and demonstrate improvements in pose and expression similarity. We also show qualitatively that our model works on objects other than faces, similarly to \cite{siarohinFirstOrderMotion2019}. Finally, we analyze the robustness of our reenactment model with respect to the feature point locations. The results indicate that our model tolerates imperfections significantly better compared to previous works.

\section{Related Work} \label{Related Work}

Face reenactment has seen a lot of interest from the research community in
the last few years and it led to photo-realistic talking heads in
\cite{zakharovFewShotAdversarialLearning2019,
  zakharovFastBilayerNeural2020a,
  wangHighResolutionImageSynthesis2018, wangFewshotVideotoVideoSynthesis, suwajanakornSynthesizingObamaLearning2017, nirkinFSGANSubjectAgnostic2019} where the source and driving
identities are the same. Representation of the pose and emotion from
the driving images is a key step to achieve higher quality talking
faces and landmarks are used as that representation in these cases. However, the landmarks
are person-specific and transfers the identity information along with the
pose and expression which leads to poor cross identity
reenactment. To address this MarioNETte
\cite{haMarioNETteFewshotFace2019, zhangFReeNetMultiIdentityFace2020}
 uses a landmark
transformer to remove person-specific information but it requires
separate hand-crafted data and model design. A few other models  \cite{tripathyICfaceInterpretableControllable2020,
  tripathyFACEGANFacialAttribute2020,
  pumarolaGANimationAnatomicallyawareFacial2018} use action units for
reenactment as they are not person-specific. However, obtaining
action units and generating photorealistic images from them with
varying head poses are challenging tasks.

In X2Face\cite{wilesX2FaceNetworkControlling2018}, as an alternative
to the landmark-based models, the pose and
expression are unsupervisedly learned from the driving images in the
form of latent codes of an encoder-decoder architecture. This approach alllows
to use other modalities, like audio, as the driver for the reenactment
models. However, these latent codes are difficult to be disentangled
from identity-like landmarks and suffer in the cross reenactment
case. In \cite{siarohinFirstOrderMotion2019,
  siarohinAnimatingArbitraryObjects2019, jakabUnsupervisedLearningObject}, similar unsupervised
learning is used to obtain keypoints from the driving and source
images from which the motion can be obtained. However, these keypoints
are similar to the landmarks and don't perform well in the
cross-reenactment case. In \cite{siarohinFirstOrderMotion2019}, they proposed to utilize
the difference of keypoints between two consecutive driving frames
as the motion cue for the source. Although it cancels out the shape of
the driving face, it only works if the source has the
same pose and expression as the first frame of the driving
video. Unfortunately, such conditions put a restriction on the choice
of source and driving pairs for the reenactment. A similar unsupervised model is
proposed in \cite{burkovNeuralHeadReenactment2020} that uses few-shot
learning at the inference time to train the model for each identity
and requires several images of the source to have better
quality results.

\begin{figure*}[t]
  \centering
  \includegraphics[scale=0.26]{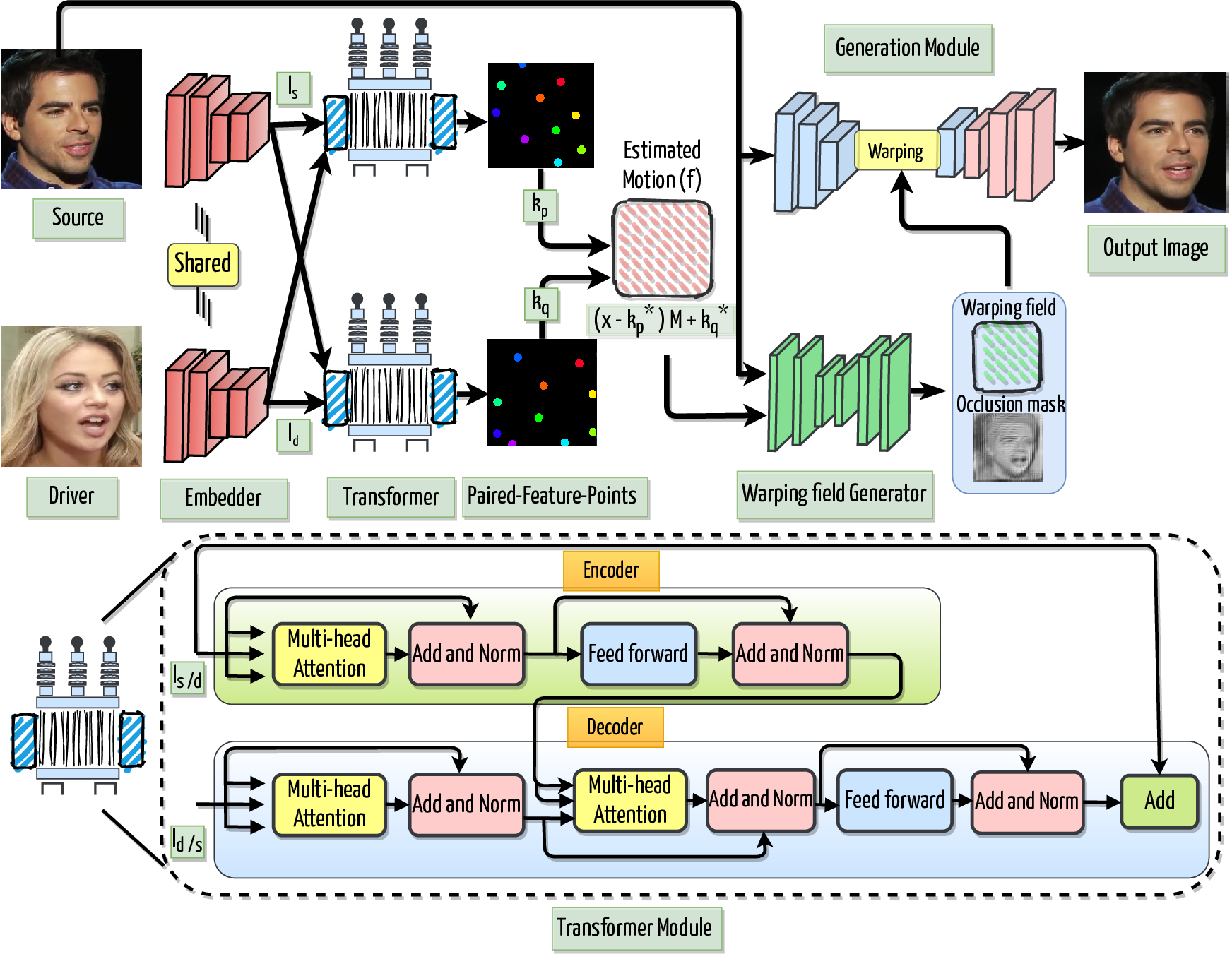}
  \caption{\label{fig:model} The complete block diagram of the
    proposed reenactment model. It processes the source and driving images in five steps as 1. Encoding the images using image embedder, 2. Extracting paired-feature-points using a transformer,
    3. Estimating the motion from paired-feature-points, 4. Converting the motion to the warping field and 5. Using the source image with the warping field in the generator to produce the final output. The transformer module is expanded at the bottom to showcase its building blocks in detail.}
\end{figure*}

Apart from the data-driven models, face reenactment has also been
performed using classical 3D face models like 3DMM
\cite{blanzMorphableModelSynthesis1999} in
\cite{thiesFace2FaceRealtimeFace}. A combination of both 3D models and
learning-based models are also used as in
\cite{kimDeepVideoPortraits2018, yaoMeshGuidedOneshot2020,
  nirkinFaceSegmentationFace2017} to create talking heads. These
reenactment models require 3D model parameters as training data
which is expensive and limits its application to a larger number of identities.


\section{Method}

The high-level architecture of our reenactment model is inspired by \cite{siarohinFirstOrderMotion2019}. That is, we first use encoded images to predict the paired feature points and the dense warping field. The warping field is subsequently applied to the source features, which are then used to generate the output image. The most important differences to \cite{siarohinFirstOrderMotion2019} are the following: 1) the motion is represented using paired-feature-points instead of keypoints extracted from individual images, and 2) the dense warping field is constructed by weighted dense motion module where the motion for each pixel is estimated by considering all paired feature points at once. These components along with the complete model are presented in the following subsections.

\subsection{Overview of our model}

Given a face image \(I_s\) of a source identity \(S\), our model aims to
animate it by copying the facial motion from a driver image
\(I_d\) with identity \(D\). The animated image generated from our model is called as
reenacted image \(I_r\).
This process of animation involves two
major steps: 1. representation of the motion difference between the
source and the driving faces, and 2. applying this motion
on to the source and creating a photo-realistic
animation of it. In our model, the representation of this motion is
obtained by understanding the motion of the feature points from the
driving
to the corresponding feature points of the source
images. Then we use this motion to backward warp the source face in the
feature space and reconstruct it using a generator with adversarial
loss.

The complete block diagram of our model is presented in Figure
\ref{fig:model}. The overall steps of our models can be summarized as,
1. representing the source and driving images as a latent vector
using an embedder network,
2. extracting the motion features by combining
both the latent codes using attention mechanism, 3. estimating pixel-wise motion
using a weighted point transformation, 4. creating a warping field from motion
using an
encoder-decoder architecture, and 5. finally reenacting the image
using an occlusion-aware generator network.

\subsection{Image Embedder}

The first block of our model is an image embedder that is used to transform \(I_s\) and \(I_d\) to latent vector representations. A single encoder network
is applied to both the \(I_s\) and \(I_d\)
independently to map them to a common space. The goal is to use the
global representations for each image that has only relevant information in obtaining the motion
points. It has a series of convolution, batch norm, and average pooling layers to
embed the images into vectors \(l_s \in \mathbb{R}^{N \times 1024}\)
and \(l_d \in \mathbb{R}^{N \times 1024}\) where \(N\) is the number
of user-defined motion features. We use \(N=10\) in all our
experiments.

\subsection{Paired-feature-point Estimation Module}

Our aim is to learn feature points from \(l_s\) and \(l_d\) that
can be used to express the motion between \(I_s\) and \(I_d\). The
state-of-the-art models like \cite{siarohinFirstOrderMotion2019,
  siarohinAnimatingArbitraryObjects2019} extract landmarks such as
\(N\) keypoints from each image individually to capture the
structure of each object (the face region in our case).
Then the motion is expressed as a function of the changes in the
structures between the source and driving images. The
model is trained to predict the structure from each
images independently without considering
the connection between the source and driving pairs as in the
reenactment.

Our motivation is to change the
feature points depending on the particularities of the \(I_s\) and
\(I_d\) together rather than considering them independently.
For example, if one of the faces in the source-driving pair has occlusion
then the feature points for the motion should be adjusted in both the
images rather than predicting keypoints in individual faces which are
highly erroneous in the occluded images. The module is implemented
using a transformer network \(T\) that maps
\(T(l_s, l_d)\) and \(T(l_d, l_s)\) to embedding vectors \(l_{st} \in
\mathbb{R}^{N \times 1024}\) and \(l_{dt} \in \mathbb{R}^{N \times
  1024}\) respectively. In reality, the transformer predicts the
changes in the \(l_s\) and \(l_t\) such that
\begin{align}
  \label{}
  l_{st} = l_s + T(l_s, l_d), \hspace{5pt} l_{dt} = l_d + T(l_d, l_s)
\end{align}
The transformer network consists of one
layer of encoder and decoder as shown in the bottom half of the
Figure \ref{fig:model}. Each of the encoder and decoder layer consists of a self
attention layer and position-wise feed forward network to map the
latent codes to an intermediate representation. The decoder consists
of an additional attention layer that is responsible to combine the
intermediate representations from both the latent codes to predict the
final embedding vector. As an example, the prediction of \(l_{st}\)
involves the self-attention of \(l_s\) from encoder, the
self attention of  \(l_{d}\) from decoder and additional co-attention
on the output of the encoder to  the output of first two layers of
decoder. Due to this co-attention layer, each embedding vector \(l_{st}\) or
\(l_{dt}\) is predicted by utilizing the latent codes \(l_s\) and
\(l_d\) from both the images.
For all the attention layers we have used a
scaled-dot-product attention (\(A\)) \cite{vaswaniAttentionAllYoub} where for
the given query (\(Q\)), key (\(K\)) and value (\(V\)), the \(A\) can
be written as,
\begin{equation}
  \label{eq:attention}
A(Q,K,V) = softmax(\frac{P(Q)P(K)^T}{\sqrt(d_k)})P(V)
\end{equation}
\(P\) is the function to calculate the sinusoidal positional encoding of the
embedding vectors. For the self-attention layer, the respective latent code
\(l_s\) or \(l_d\) are reused as the Q, K, and V pairs whereas for the
co-attention, the output at the end of the second layer of the decoder is
considered as Q and the output of the encoder is reused as K and V. We extended this attention layer
to the multi-headed attention by projecting the key, query, and value 4 times
with different feed-forward networks. We concatenate the attention from
each of them to jointly attend the information from different
projections as suggested in \cite{vaswaniAttentionAllYoub}.

\begin{figure}[t]
  \centering
  \includegraphics[scale=0.42]{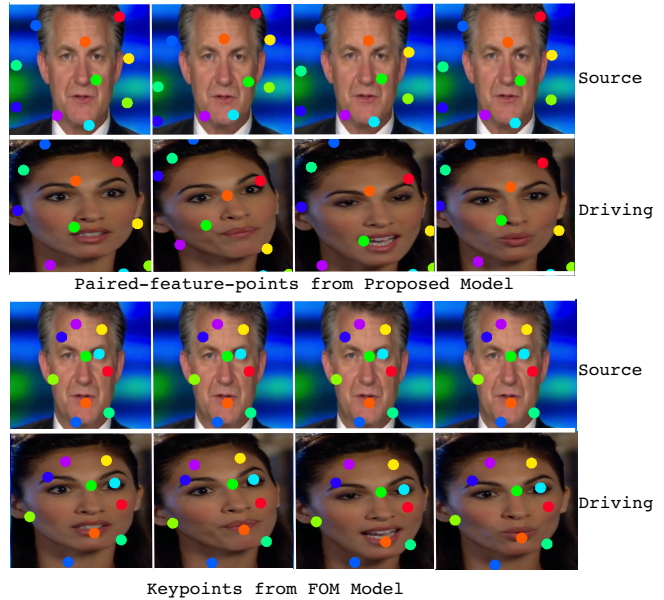}
  \caption{\label{fig:keypoint_compa} Illustration of keypoints
    predicted from FOM\cite{siarohinFirstOrderMotion2019} and
    paired-feature-points predicted from our model. During the
    reenactment, the position of the paired-feature-points are adjusted
    on both source and driving images depending on the pose and
    expression of driving image. In
    FOM\cite{siarohinFirstOrderMotion2019}, the points are
    predicted independently so source keypoints remain fixed
    throughout the process.}
  \vspace{-0.4cm}
\end{figure}

Finally, we reshape the predicted embeddings into \(\mathbb{R}^{N
\times 32 \times 32}\) and pass through a softmax layer to obtain \(N\)
heat-maps which are then converted to \(N\) points as \(k_s\) and \(k_d\)
by extracting the mean of the heat-maps. This conversion of the embeddings to \(N\) points
serve as a bottleneck and provides essential feature points for the
motion calculation. The feature points learned from our model and
keypoints from FOM
\cite{siarohinFirstOrderMotion2019} are shown in Figure \ref{fig:keypoint_compa}. It
is clear that our model does not imitate the landmark points
like FOM \cite{siarohinFirstOrderMotion2019}. One interesting aspect to note is that our source landmarks
adjust themselves according to the driving image unlike FOM \cite{siarohinFirstOrderMotion2019}
where they are fixed. We note that in
\cite{haMarioNETteFewshotFace2019} the attention layers are utilized
for reenactment but they are used to draw a spatial correspondence between
source features and driving features to effectively transfer the style
of source on to the driver's pose. Our motivation and
the design structure of the attention layers are
completely different from their counterparts. Our transformer
architecture has similarities with the point cloud registration model in \cite{wangDeepClosestPoint2019}. However, our model is designed for fundamentally different data and tasks.

\subsection{Motion Estimation}

Given the discrete motion points \(k_s\) and \(k_d\) on the \(I_s\) and
 \(I_d\) respectively, our goal is to predict the transformation \(f\)
 per each pixel point \(x\) in the driving image such that  it minimises
\begin{align}
  \label{least_square}
  \sum_n w_n \lvert f_x(k_{dn}) - k_{sn} \rvert^2
\end{align}
where for a constant \(\alpha \leq 1\), the weight \(w_n\) has the form
\begin{align}
  \label{weight}
  w_n = \frac{1}{\lvert k_{dn} - x \rvert^{2\alpha}}
\end{align}
The \(n\) is a counter on number of feature points and ranges
from 1 to \(N\). The equations \eqref{least_square} and \eqref{weight}
together constitutes the Moving Least Squares (MLS)
\cite{schaeferImageDeformationUsing} formulation where each point
\(x\) has its own transformation \(f_x\) depending the on its distance
from all other feature points. Following the derivation from
\cite{schaeferImageDeformationUsing} the final form of the \(f\) is,
\begin{align}
  \label{final_motion}
f_x = (x - k_d^*)M + k_s^*
\end{align}
where \(M\) is the linear transformation matrix, \(k_d^*=\frac{\sum
  w_nk_{dn}}{\sum w_n}\) and \(k_s^*=\frac{\sum
  w_nk_{sn}}{\sum w_n}\). In \cite{schaeferImageDeformationUsing}, the
feature points are handpicked on the edges of the object. By
considering different classes of transformation matrix \(M\) (affine, rigid and
similarity transformation) a closed-form solution can be derived for
\(f_x\) which gives reasonable deformations. In our case, the
feature points are self-supervised and applying this closed-form,
the solution can drive the feature point detector to learn the points similar to landmarks.
Moreover, in the initial iterations, this closed-form solution can completely deform the images as the keypoints are not
stable and can break the training of the whole network. To avoid such
problems,
we employed a single convolution
layer to predict the transformation matrix from the weight matrix
\(w\) and the heatmap
representations of the \(k_s\) and \(k_d\).
We extract a single \(f_x\) for the whole image rather than
learning multiple \(f_x\) for each keypoints individually like FOM
\cite{siarohinFirstOrderMotion2019}. Our pixel-level transformations
are much stable due to the moving weight matrix unlike the
\cite{siarohinFirstOrderMotion2019} where each pixel's motion is
highly dependant on the nearest keypoints motion. Experimentally we
have shown that our motion model handles small errors in the prediction
of feature points, unlike the counterparts.

\subsection{Generation Module}
After predicting the motion \(f\), we utilized this to predict a warping
function that realigns the source features in the generator
network. To detect the occluded parts of the source face an occlusion
mask is predicted during this process to indicate in-painting
region for the generator.
Along with the knowledge of the occlusion, the realigned features are finally converted to the
reenacted source images by the generator. The warping function is
generated by a U-Net architecture which is similar to the local
motion aggregation block of FOM \cite{siarohinFirstOrderMotion2019}
but we only consider two motions i.e for \(f\) and the background
instead of \(N+1\) motions in FOM. 
\begin{figure*}[ht]
  \centering
  \includegraphics[scale=0.6]{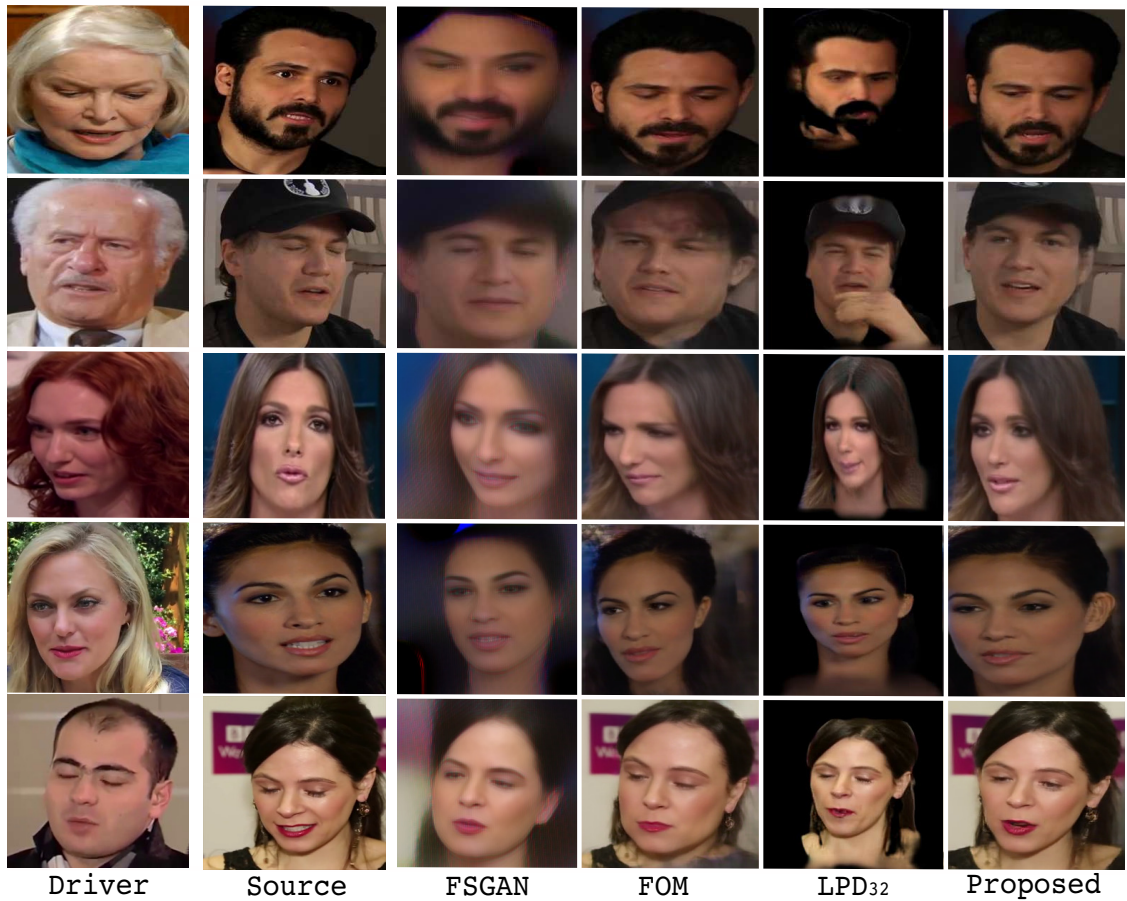}
  \caption{\label{fig:main_compa} Qualitative comparision of proposed model with FSGAN\cite{nirkinFSGANSubjectAgnostic2019}, FOM\cite{siarohinFirstOrderMotion2019} and LPD\cite{burkovNeuralHeadReenactment2020}. Our model better reproduces the source identity, facial shape and driving motion at the output. More results can be seen in the supplementary material.}
\end{figure*}

\subsection{Training Loss}
We generate the training material using videos of moving objects (faces and other shapes). We randomly sample source and driving pairs from each video, which enables us to use the driving frame as pixel-wise ground truth of the intended animation of the source image. We train our model
end-to-end using the perceptual loss
\cite{wangHighResolutionImageSynthesis2018} in multiple resolution of
\(I_r\) and \(I_d\). The mathematical expression for perceptual loss
in each resolution can be written as
\begin{align}
  \label{precep_adv}
 \mathcal{L}_{p} = \sum_i \vert \vert VGG_i(I_r) -
  VGG_i(I_d) \vert \vert_1
\end{align}
Where \(VGG_i(.)\) stands for \(i^{th}\) channel response of
pretrained VGG-19 network.

Along with the perceptual loss, we apply a loss function
to the transformation function \(f\). We apply equation
\eqref{final_motion} to equation \eqref{least_square} and after
simplifying we express our loss function as
\begin{align}
  \label{precep_adv}
 \mathcal{L}_{m} = \sum_n w_n \vert \hat{k_{dn}}M - \hat{k_{sn}} \vert^2
\end{align}
where \(\hat{k_{dn}} = k_{dn}- k_d^*\) and \(\hat{k_{sn}} = k_{sn}-
k_s^*\). The loss \(\mathcal{L}_m\) helps in predicting the transformation matrix
\(M\) which inturn predicts the per pixel motion function
\(f_x\).
In order to encourage the paired-feature-points to be spread out in the
image, a feature point spreading loss \(\mathcal{L}_f\) is applied
where the distances between the feature-points are penalised if they fall
below a threshold value. Finally the adversarial loss \(\mathcal{L}_{adv}\) is applied to
the output image to maintain the photo realism of the reenacted
image. The final loss function can be written as an weighted sum
\begin{align}
  \label{final_loss}
\mathcal{L}_{total} = \lambda_p \mathcal{L}_p + \lambda_m \mathcal{L}_m +
  \lambda_f \mathcal{L}_f +\lambda_{adv} \mathcal{L}_{adv}
\end{align}
\section{Experiments}
In this section, we assess our model in the face and non-face reenactment tasks. Moreover, we evaluate the robustness of the reenactment models with respect to the feature point (or keypoint) locations. We compare the proposed approach with the following recent works:
\begin{itemize}[leftmargin=*]
  \itemsep-0.20em
\item \textit{FSGAN} \cite{nirkinFSGANSubjectAgnostic2019} uses direct landmarks from the driving image to reenact the source face from a single image.
\item \textit{FACEGAN} \cite{tripathyFACEGANFacialAttribute2020} is a one-shot reenactment model that utilises a combination of source landmarks and driving AUs to reenact the source image.
\item \textit{FOM} \cite{siarohinFirstOrderMotion2019} learns keypoints in
  a self-supervised fashion from the source and driving images. The reenactment is done using the motion extracted from these keypoints.
\item \textit{LPD} \cite{burkovNeuralHeadReenactment2020} learns identity
  and pose descriptors in a self-supervised way from
  videos and then combines identity of source and pose
  descriptors of driving image for the reenactment. They use a few-shot learning framework to finetune the model for each source identity at the test phase. Although this is different from other methods, which are identity agnostic, we include this method as a reference to our experiments. Furthermore, we use LPD with one and 32 samples of the source identity.
\end{itemize}

The baselines cover various popular features and techniques such as
landmarks, AUs, keypoints, pose-descriptors, and few-shot
learnings presented in the reenactment literature. We note that our model uses only one source image at the test phase and generates output without any person-specific finetuning (unlike few-shot works). All the output images have a final resolution of \(256 \times 256\).




\begin{table}[t]
  \small
\begin{tabular}{p{1.5cm} |p{1.4cm} | p{0.7cm} | p{0.7cm} | p{0.7cm} | p{0.6cm}}
\hline
  Model &  Test-phase  & ISIM$\uparrow$ & PSIM$\uparrow$ & ESIM$\uparrow$ & FID$\downarrow$ \\
  & training & &  &  & \\

\hline
  \(LPD_{32}\) \cite{burkovNeuralHeadReenactment2020} & \checkmark & 0.80  & 0.67
                                         & 0.92  & - \\
  \hline
\(LPD_1\) \cite{burkovNeuralHeadReenactment2020} & \checkmark & 0.78  & 0.61 & 0.92 & - \\
  \hline
  \hline
 FSGAN \cite{nirkinFSGANSubjectAgnostic2019}  & \(\times\) & 0.39 & 0.78  & 0.91  &  192.01\\
\hline
FACEGAN \cite{tripathyFACEGANFacialAttribute2020} & \(\times\) & 0.49 & 0.84  & 0.88  &  198.75\\
\hline
FOM \cite{siarohinFirstOrderMotion2019}  & \(\times\) & 0.57  & 0.90  & 0.93  &
                                                                   127.79\\
\hline
  Proposed  & \(\times\) & 0.70  & 0.80  & 0.93  & 115.54 \\
  \hline
\end{tabular}
\caption{Quantitative comparison of our model with the state-of-the-arts. The FID scores of \(LPD\) models \cite{burkovNeuralHeadReenactment2020} are not calculated becasue they generate faces without the background unlike other models.}
\vspace{-0.4cm}
\label{table-comparision}
\end{table}

\subsection{Face-reenactment}
The face reenactment models are trained using talking head videos from Voxceleb \cite{nagraniVoxCelebLargeScaleSpeaker2017}. All the videos are preprocessed as in \cite{siarohinFirstOrderMotion2019} to obtain the source and driving frames at a resolution of \(256 \times 256\). For the evaluation, we randomly sampled 40 identities (different from those in training) from Voxceleb and FaceForensic++ \cite{rosslerFaceForensicsLearningDetect2019} datasets, and generated 80k reenacted images by taking the source and driving as different identity (cross-person setting).\

\textbf{Qualitative comparison} of our model with its
counterparts are shown in the cross-person setting in the Figure
\ref{fig:main_compa}. It is clear that the landmark and keypoint-based models like FSGAN \cite{nirkinFSGANSubjectAgnostic2019} and FOM
\cite{siarohinFirstOrderMotion2019} leak the driving facial structure
to the source face in the final image. It makes the reenacted image lose the source identity. In LPD
\cite{burkovNeuralHeadReenactment2020}, there is no shape leaking between
source and driving but they fail to replicate the facial motion
effectively as can be seen from Figure \ref{fig:main_compa} (row 1 and
2). Moreover, they require multiple source images and a source-specific training step to generate good quality final images.
In our method, the paired-feature-points are less sensitive to
facial structure, and together with our motion model,
it reproduces the source
identity with driving motion at the output better than its counterparts. Additional qualitative examples are provided in the supplementary material.\

\textbf{Quantitative comparison} of reenactment model is difficult in the cross-person setting due to lack of the exact ground truth. Nevertheless, several indirect measurements have been applied in the reenactment literature. In our experiments, we use the following metrics:
\begin{itemize}[leftmargin=*]
  \itemsep-0.4em
\item \textit{Identity Cosine Similarity between IMage embeddings (ISIM)}: It measures
  the identity similarities between source and reenacted faces by
  comparing the embeddings vectors from a pretrained face recognition
  network\cite{dengArcFaceAdditiveAngular2019a}. 
  The higher ISIM score signifies
  better identity reproduction ability at the output.
\item \textit{Pose Cosine Similarities between IMages (PSIM)}: It measures the
  cosine similarity of head pose angles of driving and reenacted faces using a pretrained
  model \cite{baltrusaitisOpenfaceFacialBehavior2018}
\item \textit{Expression Cosine Similarities between IMages (ESIM)}: To measure the expression
  retention capability of the model, the embedding vectors from a
  pretrained action unit detector \cite{liSelfSupervisedRepresentationLearning2019} is used for
  the similarity calculation.
\item \textit{Frechet-inception distance (FID)}: It measures the perceptual
  similarities of the generated images and training images.
  A lower score signifies better photo-realism of the
  reenacted images.
\end{itemize}
\begin{figure*}[ht]
  \centering
  \includegraphics[scale=0.42]{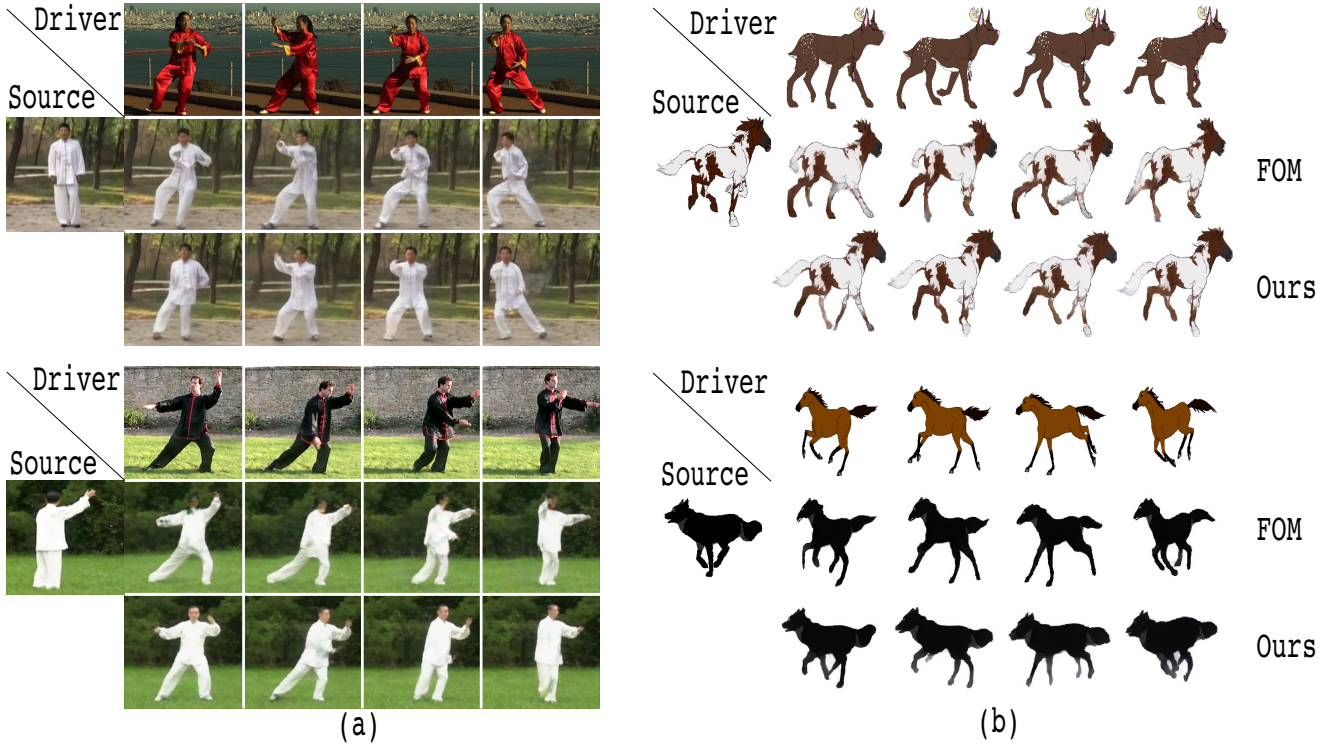}
  \caption{\label{fig:mgif} Qualitative comparison of proposed model with FOM\cite{siarohinFirstOrderMotion2019} on a. Tai-chi-HD \cite{siarohinFirstOrderMotion2019}, and b. MGif\cite{siarohinAnimatingArbitraryObjects2019} datasets. Our model keeps the source shape and driver's
 motion intact at the output unlike
 FOM\cite{siarohinFirstOrderMotion2019}. More results can be seen in
 the supplementary material.}
\vspace{-0.2cm}
\end{figure*}
The quantitative comparisons are shown in Table \ref{table-comparision}. The proposed approach achieves the highest ISIM score among all one-shot models, which illustrates the ability to retain the source identity at the output. One of the key reasons behind the performance is the pairwise-feature-points, which are shape independent unlike the keypoints in FOM
\cite{siarohinFirstOrderMotion2019} and FSGAN
\cite{nirkinFSGANSubjectAgnostic2019}. Only the \(LPD\)
models \cite{burkovNeuralHeadReenactment2020} achieve higher ISIM, which is understandable as they are trained with source identity at the test phase. The pose vectors for PSIM scores are calculated using an external pretrained network. The network utilises landmarks to obtain the poses, which makes it highly sensitive to facial contours.
In terms of PSIM, FOM \cite{siarohinFirstOrderMotion2019} and FACEGAN \cite{tripathyFACEGANFacialAttribute2020} achieve higher scores as they aim to match the landmarks between the output and the driving images. However, in this process, they hamper the image quality and identity as can be seen from the other metrics. Our model achieves a better balance between identity, pose, expression, and image quality in comparison to other models with a single source image.
\begin{table}[htbp]
\centering
\begin{tabular}{p{1.6cm} | p{2.2cm} | p{1.6cm} | p{1.3cm}}
\hline
Model &  Without Noise & With Noise & Change \\
\hline
FOM \cite{siarohinFirstOrderMotion2019}  & 1.11  & 2.19  & 97.3\% \\
\hline
Proposed  & 1.31 & 1.61 & 22.9\% \\
\hline
\end{tabular}
\caption{Mean landmark difference scores in self-reenactment between FOM
  \cite{siarohinFirstOrderMotion2019} and our model. Uniform noise is added to
a single keypoint of driving images to analyze the stability of feature-points of our model}
\vspace{-0.4cm}
\label{table-noise}
\end{table}

\subsection{Stability of paired-feature-points}
We have argued and qualitatively shown in Figure
\ref{fig:keypoint_compa} that the proposed paired feature points are different from keypoints used in FOM \cite{siarohinFirstOrderMotion2019}. To assess this, we randomly select one feature point (or keypoint) from each driving image and added uniform random noise between 0.05 to 0.5 to its location before the reenactment (point locations normalized between 0 to 1). We hypothesize that if the keypoints encode landmarks like facial structures then any distortion to it will severely distort the final image. To verify that we perform
a self-reenactment experiment using 30 identities from the test set and calculate
the mean landmark difference between output images and the
driving images as shown in Table \ref{table-noise}. The \(97\%\)
increase in the landmark error shows that FOM \cite{siarohinFirstOrderMotion2019} is
highly dependant on the correctness of keypoints and
less robust than our paired-feature-points in
the reenactment tasks. We provide qualitative examples of the output images for both methods in the supplementary material.


\subsection{Reenacting non-face objects}
The proposed formulation does not make any assumptions on the reenacted object type. Therefore, the same model can be also trained without modifications to reenact other objects besides faces. To this end, we train our method using MGif \cite{siarohinAnimatingArbitraryObjects2019} and Tai-chi-HD datasets \cite{siarohinFirstOrderMotion2019}. We provide a few qualitative reenactment examples in Figure \ref{fig:mgif}, where we compare it to FOM \cite{siarohinFirstOrderMotion2019}. The proposed model is better in preserving the source object identity compared to FOM. We provide additional examples in the supplementary material.


\section{Conclusion}
We have proposed a novel paired-feature-point detector and motion
model to unsupervisedly extract
the motion from the driver to reenact the source face. Our feature points are shape/identity
independent and represent the motion based on the source-driving pairs, unlike its
contemporaries. Our motion model predicts the motion of each source pixel
based on all the feature points instead of the closest one which makes it more stable to
any errors in feature point prediction. We have shown experimentally that our
model produces high-quality reenactment output from a single image by
keeping the desired identity, pose, expression, and photo-realism intact.


\small

\bibliographystyle{IEEEtran}

\bibliography{IEEEabrv,bibtex}

\end{document}